\documentclass{article}

\usepackage{amsmath}
\usepackage{amsfonts}
\usepackage{amssymb}
\usepackage{color}  
\usepackage{amsthm}
\usepackage{graphicx}
\usepackage{caption}
\usepackage{comment}
\usepackage{subcaption}
\usepackage{algpseudocode}
\usepackage{algorithm}
\usepackage[utf8]{inputenc}
\usepackage{tabularx}
\usepackage{multirow}
\usepackage{hyperref}

\usepackage{tcolorbox}

\DeclareMathOperator*{\argmax}{arg\,max}

\usepackage{natbib}
\bibpunct[, ]{(}{)}{,}{a}{}{,}%

\title{Multi-Robot Routing with Time Windows:\\ A Column Generation Approach}

\author{Naveed Haghani\textsuperscript{\rm 1}, Jiaoyang Li\textsuperscript{\rm 2}, Sven Koenig\textsuperscript{\rm 2},\\ Gautam Kunapuli\textsuperscript{\rm 3}, Claudio Contardo\textsuperscript{\rm 4}, Amelia Regan \textsuperscript{\rm 5}, Julian Yarkony\textsuperscript{\rm 6}\\[2ex] % All authors must be in the same font size and format. Use \Large and \textbf to achieve this result when breaking a line
\textsuperscript{\rm 1}University of Maryland, College Park, MD\\ 
\textsuperscript{\rm 2}University of Southern California, Los Angeles, CA\\ 
\textsuperscript{\rm 3}Verisk Computational and Human Intelligence Laboratory, Jersey  City, NJ\\
\textsuperscript{\rm 4}ESG UQAM and GERAD, Montreal, Canada\\
\textsuperscript{\rm 5}University of California, Irvine, CA\\
\textsuperscript{\rm 6}Laminaar, San Diego, CA
}\date{March 2021}

\begin{document}

\maketitle

\begin{abstract}

Robots performing tasks in warehouses provide the first example of wide-spread adoption of autonomous vehicles in transportation and logistics. The efficiency of these operations, which can vary widely in practice, are a key factor in the success of supply chains. In this work we consider the problem of coordinating a fleet of robots performing picking operations in a warehouse so as to maximize the net profit achieved within a time period while respecting problem- and robot-specific constraints. We formulate the problem as a weighted set packing problem where the elements in consideration are items on the warehouse floor that can be picked up and delivered within specified time windows. We enforce the constraint that robots must not collide, that each item is picked up and delivered by at most one robot, and that the number of robots active at any time does not exceed the total number available. Since the set of routes is exponential in the size of the input, we attack optimization of the resulting integer linear program using column generation, where pricing amounts to solving an elementary resource-constrained shortest-path problem. We propose an efficient optimization scheme that avoids consideration of every increment within the time windows. We also propose a heuristic pricing algorithm that can efficiently solve the pricing subproblem. While this itself is an important problem, the insights gained from solving these problems effectively can lead to new advances in other time-widow constrained vehicle routing problems.  \\ \\

\end{abstract}

\section{Introduction}
In the coming decades, adoption of autonomous vehicles including passenger cars, many different kinds of trucks, unmanned arterial vehicles (drones), and various maritime vessels will increase to the point at which autonomous operations will be the norm. However, this adoption is proceeding at a much slower pace than most experts have predicted. To date the most compelling instance of widespread adoption of autonomous vehicles in transportation and logistics is robots used in warehouses. In this paper, we tackle Multi-Robot Routing (MRR), a problem considering the challenge of efficiently routing a fleet of robots in a facility to collectively complete a set of tasks while avoiding collisions. We specifically define a problem where items are dispersed across the warehouse floor and each task involves the delivery of an item to a base location we refer to as the launcher. Items have specific pickup and delivery time windows. Robots start their routes at the launcher and must end their trip back at the launcher before a specified time limit. The number of available robots is limited and we cannot deploy more robots on the warehouse floor than are available in the fleet. We define our MRR problem as a discrete optimization problem where costs are incurred for deploying robots on the warehouse floor and for the distance they travel. Delivering an item provides a reward and the goal is to maximize the net profit (reward - cost). This amounts to optimizing the efficiency of the warehouse, not the makespan, as we expect new orders to be continuously added. The specific problem solved corresponds to an automated picking operation in which the capacity constrained robots pick up items from the warehouse and deliver these to the launcher where they are packaged for delivery. We show later that with few changes this formation can also consider the case where pallets rather than items are transported from the warehouse floor to the base station (launcher) and then returned to the warehouse floor. That variation has been described in the literature as the "Amazon" problem, though in practice it is one of many problems that large and complex warehouse operations must address. 
Our contributions are the following:

\begin{enumerate}  
\item We adapt the integer linear programming (ILP) formulation and column generation (CG) approach for a (prize collecting) vehicle routing with time windows \citep{Desrochers1992,stenger2013prize} formulation for the MRR problem.
\item We show that this formulation can incorporate important aspects of these problems that cannot be addressed in a typical Multi-Agent Path Finding (MAPF) approaches.
\item We adapt the work of \citet{boland2017continuous} to permit efficient optimization by avoiding consideration of every time increment within a window.
\item We present a heuristic for efficiently solving the elementary resource-constrained shortest-path problem (ERCSPP) during pricing. Such problems are common in a wide variety of network optimization problems so this contribution reaches beyond this application. 
\end{enumerate}

We organize this paper as follows. After a brief review of related literature in Section \ref{LitRev}, in Section \ref{setPackForm}, we formulate the MRR as an ILP, which we attack using CG in Section \ref{CGSec}. 

In Section \ref{Pricing}, we solve the corresponding pricing problem as an ERCSPP. In Section \ref{heurPRice}, we consider the use of a fast heuristic for the pricing problem with probabilistic guarantees. In Section \ref{exper}, we demonstrate the effectiveness of our approach empirically. In Section \ref{conc}, we conclude and discuss extensions.

\section{Review of Related Literature}
\label{LitRev}

Routing problems for a fleet of robots are often addressed as a Multi-Agent Path Finding (MAPF) problem \citep{SternSoCS19}. In MAPF, we are provided with a set of agents, each with an initial position and a destination, and a set of tasks that must be performed by these agents. The goal is to minimize the sum of the travel times from the initial position to the destination over all agents such that no collisions occur and all tasks are completed. MAPF can be formulated as a minimum cost multi-commodity flow problem on a space-time graph \citep{yu2013planning}. Optimization can be tackled using multiple heuristic and exact approaches, including search \citep{LiICAPS2020}, linear programming \citep{yu2013planning}, branch-cut-and-price \citep{lam2019branch}, satisfiability modulo theories \citep{SurynekIJCAI19}, and constraint programming \citep{GangeICAPS2019}.

%One key problem which is in its early stages of resolution 
One common shortcoming in typical MAPF approaches is that they require that robot task assignments be set before a robot route can be determined. The delegation of robot assignments and the creation of an optimal set of routes for the fleet are treated as independent problems. Several recent works~\citep{MaAAMAS17,LiuAAMAS19,GrenouilleauICAPS19,FarinelliAAMAS20} solve this combined problem in a hierarchical framework, i.e., assigning tasks first by ignoring the non-colliding requirement and then planning collision-free paths based on the assigned tasks. However, these assignments are sub-optimal as the consideration of possible collisions can easily affect the optimal task assignment for the fleet. Futher, MAPF approaches cannot explicitly handle time-windows which are important for routing decisions in warehouses. 

\par Our solution builds on column generation techniques for Integer Linear Programming (ILP) problems and makes use of the time-window discretization techniques presented in \citep{boland2017continuous} to reduce the number of time instances that must be explored in the huge time-space graph. Column generation is a powerful technique that repeatedly solves an LP relaxation of an ILP over a small subset of possible columns (robot routes in our problem), and because of intelligent pricing techniques arrives at an optimal solution of the original (typically intractable) problem \citep{gilmore1965multistage, barnprice, Desrochers1992, lubbecke2005selected, lubbecke2010column, zhang2017efficient, yarkony2020data}.

\par While we do not apply the elegant time window discretization methods outlined in~\citep{boland2017continuous, boland2019price} directly, the insights drawn from that work provided the motivation for our implementation. That work builds on related work such that of \citep{wang2002local, wang2009convergence} in which a time window discretization scheme is applied to solve large vehicle routing problems with time windows; the method guarantees convergence as the interval sizes approach zero (but results in much faster convergence in practice). It also draws on \citep{dash2012time}, in which the time intervals are referred to as time buckets, and in which  cutting planes which improve the solution of large-scale TSP problems with time windows are produced. All of these methods draw on the original cutting-edge ideas that were presented in \citep{appelgren1969column, appelgren1971integer} and \citep{levin1971scheduling}, decades before we had the computational power to successfully employ those ideas. 

Research on automated warehouse operations has exploded in the last few years due to advances in warehouse and robot design and operations techniques, optimization methods, meta-heuristics and hyper-heuristics for these types of problems. We mention just a few examples from the extensive literature:~\citep{sanchez2020systematic, shekari2021puzzle, weidinger2018storage, foumani2018cross}. Several recent survey papers have been published on this topic~\citep{azadeh2019robotized, boysen2019warehousing, custodio2020flexible}. These provide an excellent overview both the kinds of operations that should be considered, and the methods used to solve the problems that arise.

\section{Problem Formulation}
\label{setPackForm}
In this section, we present the Multi-Robot Routing problem and then formulate it as an ILP.  
We are given a fleet of mobile warehouse robots that enter the warehouse floor from a single location, called the launcher, pick up one or multiple items inside the warehouse, and deliver them to the launcher before the time limit. Such an operation is commonly referred to as a picker-to-goods system rather than a goods-to-picker system which is also popular. The primary difference is that in a goods-to-picker system robots would deliver whole pallets to the launcher and then return these to the warehouse (to the same or different locations) after human or robotic pickers have selected the needed items from the pallet. 
\par Each item has a reward (positive valued), and a time window during which the item can be picked up. Each robot has a capacity and is allowed to perform multiple trips. At the initial time, the fleet of robots is located at the launcher, but we also allow for some robots, called extant robots, to begin at other locations. The use of extant robots permits re-optimization as the environment changes, e.g. when rewards for picking up items change (due for example to increased urgency) or when items are added or removed. The routes for those robots begin at any location in the network and terminate at the launcher. Our goal is to plan collision-free paths for the robots to pick up and deliver items and minimize the overall cost.%, i.e., the sum of the costs for the robots to travel inside the warehouse and the costs of delivered items.

\begin{figure}[htbp!]
    \centering
	\includegraphics[width=0.70\linewidth]{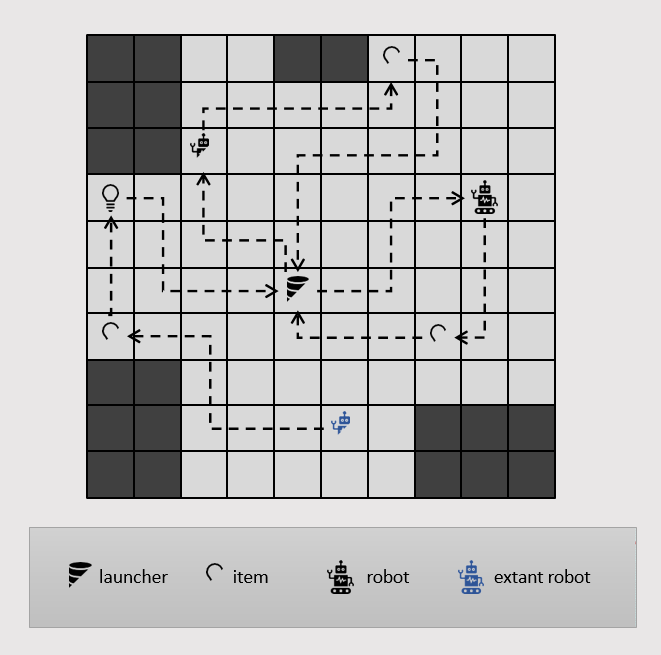}
	\caption{4-neighbor grid}
	\label{fig:Grid}
\end{figure}

For computational efficiency, we approximate the continuous space-time positions that robots occupy by treating the warehouse as a 4-neighbor grid and treating time as a set of discrete time points (See Figure \ref{fig:Grid}). This layout is fairly common, both for research and in practice \citep{shekari2021puzzle}. However, newer warehouses often have a similar grid structure, but with an additional vertical dimension (stacked pallets that can be accessed through various lifting mechanisms)~\citep{azadeh2019design}. Further many two dimensional layouts are also possible. We explore our algorithms on a 4-neighbor grid without loss of generality. These methods can be applied to more complex two- or three-dimensional warehouse operations. 
\par Each position on the grid is referred to as a cell. Cells are generally traversable, but some cells are obstructed and cannot be traversed. Through each time point, robots are capable of remaining stationary or moving to an adjacent unobstructed cell in the four main compass directions, which we connect through edges. Robots are required to avoid collisions by not occupying the same cell at any time point and not traversing the same edge in opposite directions between any successive time points. Every item is located at a unique cell. Robots incur a time based cost (negative valued) while deployed on the grid, and a distance based cost (negative valued) for moving on the grid, but obtain a reward for servicing an item (positive valued). 
\par To service an item, a robot must travel to the specific cell where the item is located during the item's associated serviceable time window and pick it up for delivery to the launcher. Servicing an item consumes a portion of the robots capacity, which is refreshed once it travels back to the launcher. The complete path a specific robot takes, which necessarily ends at the launcher, is called a route.

%We consider the problem of MRR in the context of weighted set packing.  In our formulation the goal is to minimize the objective (maximize the reward) by delivering items for processing using the robots available within a time limit and respecting the problem specific properties of the robots.  The robots must pick items and return them under the following constraints. 
%(1) An item can not be serviced (picked up and delivered) more than once.
%(2) We can use no more robots than we have available (denoted $N$).
%(3) Robots can not collide.
%(4) Each individual route must be feasible for a robot.  The set of feasible routes is a function of numerous robot-specific properties including but not limited to: battery charge, time windows for pickup/delivery of items, and the capacity of the robot. \jl{I don't think we consider battery charge in the formulation?}
%(5) There is a single location called the launcher from which robots start and drop off deliveries.  In addition, robots can be active at the start of optimization at locations other than the launcher.  Such robots are referred to as extant.  This permits re-optimization as the environment changes such as items having rewards change or items being added or removed.  

%For computational efficiency, we approximate the continuous space-time that robots occupy by treating the warehouse as a grid graph (with items and obstacles) at each of a set of discrete times and robots as taking up a single point in space at any given time when active. 
%
We formulate MRR as an ILP problem using the following notation. \\

\begin{tcolorbox}
%
%\begin{align}
%
\textbf{NOTATION}\\
\textbf{Let} $(\mathcal{P},\mathcal{E})$ \textbf{be the time-extended graph}\\
\textbf{Indices and Sets:}\\ \\
$g  \in \mathcal{G}$  feasible robot routes (too large to be enumerated)\\
$g \in \hat{\mathcal{G}}$ the subset of routes included in the restricted master problem (RMP)\\
$d  \in \mathcal{D}$  items\\
$r  \in \mathcal{R}$  extant robots\\
$t \in  \mathcal{T}$  time \\
$p  \in \mathcal{P}$  space-time position\\
$e  \in \mathcal{E}$  edges in the time-extended graph\\
$\bar{e} \in \bar{\mathcal{E}}$ conflicting edges in the time-extended graph\\ 
\textbf{Data:}\\ \\
$N$ total number of available robots\\
$\Gamma_g$ net profit of robot route $g$\\
$G_{dg}=1$ IFF route $g$ services item $d$\\
$G_{rg}=1$ IFF route $g$ is served by extant robot $r$\\
$G_{tg}=1$ IFF route $g$ is served at time $t$\\
$G_{eg}=1$ IFF route $g$ uses space-time edge $e$\\
Pickup time windows $[t^-_d,t^+_d]$ associated with each item $d \in \mathcal{D}$ \\
$\theta_d$ is the reward for picking up item $d$ \\
$\theta_1$ is the cost per unit time associated with a robot in service (moving or waiting) \\
$\theta_2$ is the cost per unit distance of moving a robot \\
$c_0 \in \mathcal{Z^+}$ the unit capacity of each robot \\
$c_d$ the number of units in item $d$\\
\textbf{Decision Variables:} \\ 
$\gamma_g=1$ IFF route $g$ is used in the solution
%
%\end{align}
%
\end{tcolorbox}

\newpage
We use $\mathcal{G}$ to denote the set of feasible robot routes, which we index by $g$. We note that $\mathcal{G}$ is too large to be enumerated. We use $\Gamma_g \in \mathbb{R}$ to denote the net profit of robot route $g$. We use $\gamma_g \in \{0,1\}$ to describe a solution which includes $g$ IFF $\gamma_g=1$. 
\par We describe the sets of items, times, and extant robots as $\mathcal{D}$, $\mathcal{T}$, and $\mathcal{R}$, respectively, which we index by $d$, $t$, and $r$, respectively.  %Here $e$ corresponds to pairs to a pair of space positions and 
\par
We use $(\mathcal{P},\mathcal{E})$ to denote the time-extended graph. Every $p \in \mathcal{P}$ represents a space-time position, which is determined by a location (i.e., an unobstructed cell on the warehouse grid) and a time $t \in \mathcal{T}$. Two space-time positions $p_i, p_j \in \mathcal{P}$ are connected by a (directed) space-time edge $e=(p_i, p_j) \in \mathcal{E}$ IFF the locations of $p_i$ and $p_j$ are the same cell or adjacent cells and the time of $p_j$ is the time of $p_i$ plus one.  % So as to prevent robot collisionsthat correspond to a swap of spatial positions at a given point in time, 
\par
We define $\bar{\mathcal{E}}$ as the set of pairs of conflicting edges, which we index as $\bar{e}=(e_1,e_2)$. A pair of space-time edges $(e_1,e_2)$ lies in $\bar{\mathcal{E}}$ IFF the transitions occur at the same point in time and between the same two points in space but in opposite directions. We use this set in our mathematical model to prevent collisions. 

%\\  %There is one  any $e_1$ % This is defined for any space-time edges $e_1=(p_1\rgih)%For example, two edges $e_1,e_2\in\mathcal{E}$ translate to the same element $\bar{e}\in\bar{\mathcal{E}}$ if $e_1$ and $e_2$ represent at time $t$ $(p_1)$ transitioning to $p_2$ and $p_2$ transitioning to $p_1$ respectively.  

We describe routes using $G_{ig} \in \{0,1\}$ for $i \in \mathcal{I}= \{ \mathcal{D} \cup \mathcal{T}\cup \mathcal{P}\cup \mathcal{E} \cup \mathcal{R} \}$.  
\par We set $G_{dg}=1$ IFF route $g$ services item $d$.  We set $G_{tg}=1$ IFF route $g$ is active (meaning moving or waiting) at time $t$. 
\par We set $G_{pg}=1$ IFF route $g$ includes space-time position $p$. We set $G_{rg}=1$ IFF route $g$ is associated with extant robot $r$.  
\par We set $G_{eg}=1$ IFF route $g$ uses a space-time edge $e$. We use $N$ to denote the total number of robots available in the fleet.  \\

We write MRR as an ILP as follows, followed by an explanation of the objective and constraints. %Below RMP refers to the Restricted Master Problem. 

\begin{align}
\label{formalOpt}
    \max_{\substack{\gamma_g  \in \{ 0,1\} \, \forall g \in \mathcal{G}}}% \xi_d \geq 0 \quad \forall d \in \mathcal{D}} }
    \sum_{g \in \mathcal{G}}\Gamma_g \gamma_g\\%-\sum_{d \in \mathcal{D}}\theta_d \xi_d \\
    \label{max_delv}
    \sum_{g \in \mathcal{G}} G_{dg}\gamma_g \leq 1\quad \forall d \in \mathcal{D} \\
    \label{max_time}
    \sum_{g \in \mathcal{G}} G_{tg}\gamma_g \leq N \quad \forall t \in \mathcal{T}  \\
    \label{max_rob}
    \sum_{g \in \mathcal{G}} G_{rg}\gamma_g = 1 \quad \forall r \in \mathcal{R} \\
    \label{max_pos}
    \sum_{g \in \mathcal{G}} G_{pg}\gamma_g \leq 1 \quad \forall p \in \mathcal{P}  \\
    \label{max_edge}
    \sum_{g \in \mathcal{G}} (G_{e_1g}+G_{e_2g})\gamma_g \leq 1 \quad \forall (e_1,e_2) \in \bar{\mathcal{E}}  
\end{align}

In \eqref{formalOpt}, we maximize the net profit (that includes the rewards collected minus the operational costs of the robots) of the MRR solution. \\
In \eqref{max_delv}, we enforce the constraints that no item is serviced more than once. \\
In \eqref{max_time}, we enforce the constraints that no more than the available number of robots $N$ are used at any given time. \\ 
In \eqref{max_rob}, we enforce that each extant robot is associated with exactly one route. \\
In \eqref{max_pos}, we enforce the constraint that no more than one robot can occupy a given space-time position. \\ 
In \eqref{max_edge}, we enforce that no more than one robot can use a space-time edge in any given conflicting set; thus preventing collisions and also preventing pairs of robots from swapping spatial positions at a given point in time.

Here we describe a set of feasibility constraints and cost terms for robot routes in our application. % However other feasibility/cost terms can be used as long as \eqref{pricing_prob} can be solved.%n our application we consider the following.  
 (a) Each item $d \in \mathcal{D}$ can only be picked up during its time window $[t^-_d,t^+_d]$. (b) Each item $d \in \mathcal{D}$ uses $c_d \in \mathbb{Z}_{+}$ units of capacity of a robot. The capacity of a robot is $c_0 \in \mathbb{Z}_{+}$. An  active (extant) robot $r \in \mathcal{R}$ is associated with an initial space-time position $p_{0r}$ (at the initial time, i.e., time 1) and a remaining capacity $c_r \in [0, c_0]$.

The cost associated with a robot route is defined by the following terms. (a) $\theta_d \in \mathbb{R}_{+}$ is the reward associated with servicing item $d$. (b) $\theta_1,\theta_2 \in \mathbb{R}_{0-}$ are the costs of being on the floor and moving respectively, which can account for both operational and depreciation costs. Using $\theta_d$, $\theta_1$, and $\theta_2$, we write $\Gamma_g$ as follows.

\begin{align}
    \Gamma_{g}=\sum_{d \in \mathcal{D}}\theta_dG_{dg}+\sum_{t \in \mathcal{T}}\theta_1G_{tg}+\sum_{e \in \mathcal{E}}\theta_{2}G_{eg}
\end{align}

\section{Column Generation for MRR}
\label{CGSec}

\par Recall that in each iteration of a column generation method, two problems must be solved. The first is the restricted master problem (RMP) and the second is the subproblem which in this application is an elementary resource constrained shortest path problem (ERCSPP). 
\par The RMP is the original problem with only a subset of variables (routes). By solving the RMP, a vector of dual values associated with the constraints of the RMP is obtained. The dual vector is passed on to the subproblem. The goal of the subproblem is to identify a new route and an associated coefficient column with negative reduced cost. Such columns have the potential to improve the objective function value of the original problem. If such a route can be identified, then it is added to RMP, which is re-optimised, and the next iteration begins. If no such routes (columns) are found, then an optimal solution of the RMP is also an optimal solution of the original problem. In practice more than one (and even many) reduced cost routes can be added to the RMP in each iteration. 

Since $\mathcal{G}$ cannot be enumerated in practice, we attack optimization in  \eqref{formalOpt}-\eqref{max_edge} using column generation (CG). Specifically, we relax $\gamma$ to be non-negative and construct a sufficient set $\hat{\mathcal{G}} \subset \mathcal{G}$ to solve optimization over $\mathcal{G}$ using CG. CG iterates between solving the LP relaxation of \eqref{formalOpt}-\eqref{max_edge} over $\hat{\mathcal{G}}$, which is referred to as the Restricted Master Problem (RMP), followed by adding elements to $\hat{\mathcal{G}}$ that have positive reduced cost, which is referred to as pricing. Below we formulate pricing as an optimization problem using $\lambda_{d}$, $\lambda_t$, $\lambda_r$, $\lambda_p$, and $\lambda_{e_1e_2}$ to refer to the dual variables over constraints \eqref{max_delv}-\eqref{max_edge} of the RMP respectively. For each $(e_1,e_2) \in \bar{\mathcal{E}}$ we define $\lambda_{e_1}\leftarrow \lambda_{e_1e_2}$ and $\lambda_{e_2}\leftarrow \lambda_{e_1e_2}$.
\begin{align}
\label{pricing_prob}
    \max_{g \in \mathcal{G}}\bar{\Gamma}_g \quad \mbox{where} \quad 
    \bar{\Gamma}_g=\Gamma_g -\sum_{i \in \mathcal{I}} \lambda_i G_{ig}  
\end{align}

We terminate optimization when the solution to \eqref{pricing_prob} is non-positive, which means that $\hat{\mathcal{G}}$ is provably sufficient to exactly solve the LP relaxation of optimization over $\mathcal{G}$~\citep{lubbecke2005selected}. 
\par We initialize $\hat{\mathcal{G}}$ with any feasible  solution (perhaps greedily constructed) so as to ensure that each $r \in \mathcal{R}$ is associated with a route. \par At termination of CG, if $\gamma_g \in \{0,1\}, \forall g \in \mathcal{G}$, then the solution, i.e. the routes defined by $\{g\in \mathcal{G} | \gamma_g=1\}$, is provably optimal.  Otherwise, an approximate solution can be produced by solving the ILP formulation over $\hat{\mathcal{G}}$ or the formulation can be tightened using valid inequalities, such as subset-row inequalities \citep{jepsen2008subset}.  We can also use branch-and-price \citep{barnprice} to formulate CG inside a branch-and-bound formulation. 

\par
\begin{figure}[!htbp]
    \centering
	\includegraphics[width=.5\linewidth]{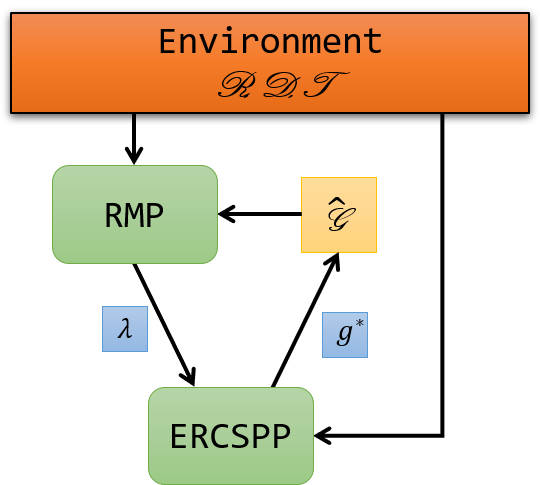}
	\caption{Visualization of the CG algorithm.\\The RMP solves the primal and delivers the dual variables ($\lambda$) to the ERCSPP pricing algorithm. The pricing algorithm solves the ERCSPP and delivers the optimal route(s) found ($g^{*}$). The set of columns $\hat{\mathcal{G}}$ is updated to include the new route(s), and the new set is inputted to the RMP to be resolved. The environment passes its relevant parameters (items, time windows, and extant robots) to the necessary modules.}
	\label{fig:CG_env}
\end{figure}

 Figure \ref{fig:CG_env} shows a visualization of the CG algorithm. Algorithm~\ref{BasicColGenAlg} shows the pseudocode for CG. We provide a discussion of an enhanced version of CG motivated by dual optimal inequalities (DOI) in section \ref{doi_sec} which follows \citep{ben2006dual}.

\begin{algorithm}[H]%[1][H]
 \caption{Optimization via Column Generation}
\begin{algorithmic}[1] 
\Repeat
\State $\gamma,\lambda \leftarrow $ Solve the RMP  over $\hat{\mathcal{G}}$
\State $g^* \leftarrow \argmax_{g \in \mathcal{G}}\bar{~\Gamma_g}$
\State $\hat{\mathcal{G}} \leftarrow \hat{\mathcal{G}} \cup \{ g^*\} $
 \Until{ $\bar{\Gamma}_{g^*} \leq 0$ }
 \State $\gamma \leftarrow$ Solve ILP in \eqref{formalOpt}-\eqref{max_edge} over $\hat{\mathcal{G}}$ instead of $\mathcal{G}$
\State Return $\gamma$
\end{algorithmic}
\label{BasicColGenAlg}
\end{algorithm}

\newpage
\section{Dual Optimal Inequalities}
\label{doi_sec}
In this section, we provide dual optimal inequalities (DOI) for MRR, which accelerate CG and motivate better approximate solutions at termination of CG when the LP relaxation is loose. Our DOI are motivated by the following observation. No optimal solution to \eqref{pricing_prob} services an item $d$ that is associated with a net penalty instead of a net reward for being serviced, meaning that $\theta_d\leq \lambda_d $ must be observed. 
This is so because simply not servicing the item but using an identical route in space and time would produce a lower reduced cost route.  

\par Since the dual LP relaxation of  \eqref{formalOpt}-\eqref{max_edge} is increasing with respect to $\lambda$, no optimal dual solution to  \eqref{formalOpt}-\eqref{max_edge} will violate $\theta_d\leq \lambda_d \quad  \forall d \in \mathcal{D}$.
By enforcing $\theta_d\leq \lambda_d \quad  \forall d \in \mathcal{D}$ at each iteration of CG optimization, we accelerate CG by restricting the dual space that needs to be explored. In the primal form, Eq \eqref{formalOpt} and Eq \eqref{max_delv} are altered as follows with primal variables $\xi_d$ corresponding to $\theta_d\leq \lambda_d \quad  \forall d \in \mathcal{D}$.

%$\eqref{formalOpt}\mbox{ becomes } \min_{\substack{\gamma_g\geq 0 \\ \xi\geq 0}}% \xi_d \geq 0 \quad \forall d \in \mathcal{D}} }
    %\sum_{g \in \mathcal{G}}\Gamma_g \gamma_g -\sum_{d \in \mathcal{D}}\theta_d \xi_d \nonumber\\
    %\mbox{and }\eqref{max_delv}\mbox{ becomes }\sum_{g \in \mathcal{G}} G_{dg}\gamma_g \leq 1+\xi_d\quad \forall d \in \mathcal{D} \nonumber$    
\begin{align}
   \eqref{formalOpt}\mbox{ becomes } \max_{\substack{\gamma_g\geq 0 \\ \xi\geq 0}}% \xi_d \geq 0 \quad \forall d \in \mathcal{D}} }
    \sum_{g \in \mathcal{G}}\Gamma_g \gamma_g -\sum_{d \in \mathcal{D}}\theta_d \xi_d \nonumber\\
    \mbox{and }\eqref{max_delv}\mbox{ becomes }\sum_{g \in \mathcal{G}} G_{dg}\gamma_g \leq 1+\xi_d\quad \forall d \in \mathcal{D} \nonumber
\end{align}

In our experiments, we use the replacements above when solving the ILP over the column set $\hat{\mathcal{G}}$. When enforcing that $\gamma$ is binary, the technique described often leads to a closer approximations to the solution to Eq \eqref{formalOpt}-\eqref{max_edge}. We map any solution derived this way to one solving the original ILP by arbitrarily removing over-included items from routes in the output solution until each item is included no more than once.

\section{Solving the Pricing Problem }
\label{Pricing}
In this section, we consider  the problem of pricing, which we show is an elementary resource-constrained shortest-path problem (ERCSPP) \citep{righini2008new}.  We organize this section as follows. In Section \ref{trivialPricing}, we  formulate pricing as an ERCSPP over a graph whose nodes correspond to space-time positions and whose resources correspond to the items to be picked up. In Section \ref{efficentPricing}, we accelerate computation from Section \ref{trivialPricing} by coarsening the graph, leaving only locations of significance (here these are locations containing item to be picked up). In Section \ref{removeExplicitTIme}, we further accelerate computation by limiting the times considered while still achieving exact optimization during pricing. In Section \ref{partOptFaster}, we show that CG can be accelerated by updating the $\lambda_i$ for all $i \in \mathcal{D}\cup \mathcal{R}$ more often than the remainder of the dual solution, saving computation time by eliminating the need to reconstruct the coarsened graph between each round of pricing. 
\subsection{Basic Pricing}
\label{trivialPricing}
In this section we establish a weighted graph admitting an injunction from the routes in $\mathcal{G}$ to the paths in the graph. By that we mean that every route corresponds to at most one path, but paths do not necessarily correspond to routes (and in fact most do not). For a given route $g$, the sum of the weights along the corresponding path in the weighted graph is equal to the route's reduced cost $\bar{\Gamma}_g$. Thus finding the highest-profit feasible (where feasible means corresponding to a member of $\mathcal{G}$) path in this graph solves \eqref{pricing_prob}. The graph proposed is a modified form of the time-extended graph $(\mathcal{P},\mathcal{E})$. Nodes are added to represent start/end locations, item pickups, and the use of extant robots. Remember that extant robots can begin at any location in the graph and that they can have reduced carrying capacities due to items that have already been picked up. Weights are amended by the corresponding dual variables associated with a given node/edge. We solve an ERCSPP over this graph where the resources are the items to be picked up. %and their associated capacity requirements.
%In the weighted graph, we have a copy of the time-extended graph. We add a source node, a sink node and item-pickup nodes at different times in its time window that signify a pickup. % (you could drive right by the item without actually servicing it).

Formally, consider a graph $(\mathcal{P}^+,\mathcal{E}^+)$ with paths described by $x_{p_ip_jg}\in \{0,1\}$ for $(p_i,p_j) \in \mathcal{E}^+,g \in \mathcal{G}$, where $x_{p_ip_jg}=1$ indicates that edge $(p_i,p_j)$ is traversed by the path on the graph corresponding to route $g$. 

\par Each edge $(p_i,p_j)$ has an associated weight $\kappa_{p_ip_j}$. There is a node in $\mathcal{P}^+$ for each $p\in \mathcal{P}$, for each pairing of $d\in \mathcal{D}$ and $t \in [t_d^-,t_d^+]$ denoted $p_{dt}$, for each $r \in \mathcal{R}$ denoted $p_r$, the source node $p_+$, and the sink node $p_-$.  We ensure that  $\bar{\Gamma}_g=\sum_{(p_i,p_j) \in \mathcal{E}^+}\kappa_{p_ip_j}x_{p_ip_jg}$ for all $g \in \mathcal{G}$. 

\par For each pair of space-time positions $p_i,p_j$ occurring at the same cell at times $t_i,t_j=t_i+1$ (representing a wait action),  we set  $\kappa_{p_ip_j}=\theta_{1}-\lambda_{t_j}-\lambda_{p_j}$. We set $x_{p_ip_jg}=1$ IFF robot route $g$ transfers from $p_i$ to $p_j$ and no pickup is made at $p_i$.

\par For each pair of space-time positions $p_i,p_j$ occurring at times $t_i$ and $t_j=t_i+1$ and associated with space-time edge $e$ (representing a move action), we set $\kappa_{p_ip_j}=\theta_{1}+\theta_{2}-\lambda_e-\lambda_{t_j}-\lambda_{p_j}$. We set $x_{p_ip_jg}=1$ IFF robot route $g$ transfers from $p_i$ to $p_j$ and no pickup is made at $p_i$.
\par For each $d \in \mathcal{D},t\in [t_d^-,t_d^+]$, which occurs at space-time position $p$, we set $\kappa_{p p_{dt}}=\theta_d-\lambda_{d}$.  We set $x_{p p_{dt}g}=1$ IFF robot route $g$  picks up item $d$ at time $t$.  

\par For each $d \in \mathcal{D},t\in [t_d^-,t_d^+]$, which occurs at an associated $p$, we provide identical outgoing $\kappa$ terms for $p_{dt}$ as we do $p$ (except there is no self connection $p_{dt}$ to $p_{dt}$).  We set $x_{p_{dt}p_jg}=1$ IFF robot route $g$ transfers from the position of item $d$ to $p_j$ and item $d$ is picked up at time $t_j-1$ on route $g$. 
\par For each $t \in \mathcal{T}$ we connect the $p_+$ to the launcher at time $t$ denoted $p_{0t}$ with weight $ \kappa_{p_+p_{0t}}=\theta_1-\lambda_{t}-\lambda_{p_{0t}}$.  We set $x_{p_+p_{0t}g}=1$ IFF the robot route $g$ appears first at  $p_{0t}$.  
\par For each $r \in \mathcal{R}$  we set $\kappa_{p_+p_{r}}=\theta_1-\lambda_r-\lambda_{t=1}-\lambda_{p_r} $. We set $x_{p_+p_{r}g}=1$ IFF the robot route $g$ appears first at $p_{r}$.  For each $r \in \mathcal{R}$, $p_r$ has one single outgoing connection to $p_{0r}$ with weight $\kappa_{p_rp_{0r}}=0$.

%For each $r \in \mathcal{R}$ we set $\kappa_{p_+p_{r}}=\theta_1-\lambda_r$. If $p$ is the corresponding start position for extant robot $r$ at $t=1$, we set $\kappa_{p_rp}=-\lambda_{t=1}-\lambda_{p}$.  We set $x_{p_+p_{r}g}=1$ IFF the robot route $g$ corresponds to extant robot $r$.

\par For each $t \in \mathcal{T}$ we set $\kappa_{p_{0t}p_-}=0$. We set $x_{p_{0t}p_-g}=1$ IFF the robot route $g$ has its last position at  $p_{0t}$.
\par Using $\kappa$ defined above we express the solution to \eqref{pricing_prob} as an ILP ( followed by description) using decision variables $x_{p_ip_j}\in \{0,1\}$ where $x_{p_ip_j}$ is equal to $x_{p_ip_jg}$ for all $(p_i,p_j) \in \mathcal{E}^+$.

\begin{align}
\label{objPath}
    \max_{x_{p_ip_j} \in \{0,1\} \quad \forall (p_i,p_j) \in \mathcal{E}^+}\sum_{(p_i,p_j) \in \mathcal{E}^+}\kappa_{p_ip_j}x_{p_ip_j}\\
   % \sum_{(p_+,p) \in \mathcal{E}^+} x_{p_+p}=[]\\
    %\sum_{(p_+,p) \in \mathcal{E}^+} x_{pp_-}=1\\
    \label{flowConst}
    \sum_{(p_i,p_j) \in \mathcal{E}^+} x_{p_ip_j}-\sum_{(p_j,p_i) \in \mathcal{E}^+} x_{p_jp_i}\\
    \nonumber =[p_i=p_+]-[p_i=p_-] \quad \forall p_i \in \mathcal{P}^+  \\
    \label{capacityConst}
    \sum_{d \in \mathcal{D}} c_d \sum_{t_{d}^-\leq t\leq t_{d}^+} \sum_{(p,p_{dt}) \in \mathcal{E}^+} x_{pp_{dt}} \leq c_0+\sum_{r \in \mathcal{R}}(c_r-c_0)x_{p_+p_r}  \\
    \label{resourceConst}
    \sum_{t_{d}^-\leq t\leq t_{d}^+}  \sum_{(p,p_{dt}) \in \mathcal{E}^+} x_{pp_{dt}} \leq 1 \quad \forall d \in \mathcal{D}  
\end{align}
%\jl{In (10), shouldn't $p_j$ on the rhs be $p_i$? Besides, on the lhs, I think the sum is over all incoming/outgoing edges of $p_i$, rather than all edges in $\mathcal{E}^+$. }
In \eqref{objPath} we provide objective s.t. $\bar{\Gamma}_g=\sum_{(p_i,p_j) \in \mathcal{E}^+}\kappa_{p_ip_j}x_{p_ip_jg}$ for all $g \in \mathcal{G}$. In \eqref{flowConst} we ensure that $x$ describes a path from $p_+$ to $p_-$ across space and time. In \eqref{capacityConst} we ensure that capacity is obeyed. In \eqref{resourceConst} we ensure that each item is picked up at most once.  Optimization in \eqref{objPath}-\eqref{resourceConst} is strongly NP-hard as complexity grows exponentially with $|\mathcal{D}|$ \citep{Desrochers1992}.  

\subsection{Efficient Pricing: Considering Only Nodes Corresponding to Items}
\label{efficentPricing}
In this section we rewrite the optimization for pricing in a manner that vastly decreases the size of the graph under consideration. This transformation leads to efficient optimal solutions for the ERCSPP. We exploit the fact that given the space-time positions where item pickups occur, we can solve of the remainder of the problem as independent parts. Each such independent part is solved as a simple shortest path problem, which can be solved via a shortest path algorithm such as Dijkstra's algorithm \citep{dijkstra1959note}. 

We now consider a graph with node set $\mathcal{P}^2$  with edge set $\mathcal{E}^2$, decision $x^2_{p_ip_jg} \in \{0,1\}$ and weights $\kappa^2$. There is one node in $\mathcal{P}^2$ for each $p \in \mathcal{P}^+$ excluding those for $p \in \mathcal{P}$, i.e., $\mathcal{P}^2 = \mathcal{P}^+\setminus\mathcal{P}$.  
For any $p_i,p_j \in \mathcal{P}^2$, $(p_i,p_j) \in \mathcal{E}^2$ IFF there exists a path from $p_i$ to $p_j$ in $\mathcal{E}^+$ traversing only intermediate nodes that exist in $\mathcal{P}$. We set $\kappa^2_{p_ip_j}$ to be the weight of the shortest path from $p_i$ to $p_j$ in $\mathcal{E}^+$ using only intermediate nodes in $\mathcal{P}$. This is easily computed using a shortest path algorithm. We set $x^2_{p_ip_jg}=1$ IFF  $p_i$ is followed by $p_j$ in robot route $g$ when ignoring nodes in $\mathcal{P}$.  Replacing $\mathcal{E}^+,x$ with $\mathcal{E}^2,x^2$ respectively in \eqref{objPath}-\eqref{resourceConst} we have a smaller but equivalent optimization problem permitting more efficient optimization.

%\jl{If I understand correctly, 
%the size of $\mathcal{P}^+$ is $|\mathcal{P}|^+ = 2 + |\mathcal{R}| + \sum_{d \in \mathcal{D}}(t_{d}^+ + 1 - t_{d}^-) + |\mathcal{P}|$,  
%the size of $\mathcal{E}^+$ is $|\mathcal{E}^+| = 2 + |\mathcal{R}| + 2\sum_{d \in \mathcal{D}}(t_{d}^+ + 1 - t_{d}^-) + |\mathcal{E}|$,
%the size of $\mathcal{P}^2$ is $|\mathcal{P}^2| = 2 + |\mathcal{R}| + \sum_{d \in \mathcal{D}}(t_{d}^+ + 1 - t_{d}^-)$,
%and the size of $\mathcal{E}^2$ is $|\mathcal{E}^2| \leq 2|\mathcal{R}| + |\mathcal{R}|\sum_{d \in \mathcal{D}}(t_{d}^+ + 1 - t_{d}^-) + \sum_{d_1, d_2 \in \mathcal{D}, d_1 \neq d_2}(t_{d_1}^+ + 1 - t_{d_1}^-)(t_{d_2}^+ + 1 - t_{d_2}^-) + 2\sum_{d \in \mathcal{D}}(t_{d}^+ + 1 - t_{d}^-)$ (i.e., edges between $p_r$ and $p_+$/$p_-$ plus edges between $p_r$ and $p_{dt}$ plus edges between $p_{dt}$ and $p_{dt}$ plus edges between $p_{dt}$ and $p_+$/$p_-$.
%I want to confirm this because I suggest we compute the size of the graphs in Sections 4.1, 4.2 and 4.3 respectively so that we can directly show how many variables we can save by using efficient models.}

%
\subsection{More Efficient Pricing: Avoiding Explicit Consideration of All Time Instances}
\label{removeExplicitTIme}
The optimization in Eq \eqref{objPath}-\eqref{resourceConst} over $\mathcal{E}^2$ requires the enumeration of all $d \in \mathcal{D}, t \in [t_d^-,t_d^+]$ pairs, which is expensive. In this section we circumvent the enumeration of all $d \in \mathcal{D}, t \in [t_d^-,t_d^+]$ pairs by aggregating time into sets in such a manner so as to ensure exact optimization during pricing. Here we draw on the insights discussed in \citep{boland2017continuous} with respect to continuous-time service network design problems. 
\par For every $d \in \mathcal{D}$, we construct $\mathcal{T}_{d}$, which is an ordered subset of the times $[t_d^-,t_d^+ +1]$ where $\mathcal{T}_d$ initially includes $t_d^-$ and $t_d^+ +1$ and is augmented as needed.  \par We order these in time where $\mathcal{T}_{dj}$ is the $j$'th value ordered from earliest to latest.  \par $\mathcal{T}_d$ defines a partition of the window $[t_d^-,t_d^+]$ into $|\mathcal{T}_d|-1$ sets, where the $j$'th set is defined by $[\mathcal{T}_{dj},\mathcal{T}_{dj+1}-1]$.

We use $\mathcal{P}^3,\mathcal{E}^3,\kappa^3,x^3$ to define the graph and solution mapping. Here $\mathcal{P}^3$ consists of $p_+,p_-, p_r \forall r \in \mathcal{R}$ and one node $p_{dj}$ for each $d \in \mathcal{D},j \in \mathcal{T}_{d}$.  

\par We define $x^3_{p_+p_{dj}g}=1$ if route $g$ services item $d$ at a time in $[\mathcal{T}_{dj},\mathcal{T}_{d \; j+1}-1]$ as its first pick up.  The remaining $x$ terms are defined similarly over aggregated time sets.  

\par We assign each $\kappa^3_{p_ip_k}$ to be some maximum $\kappa^2$ over the possible paths in $(\mathcal{P}^2,\mathcal{E}^2)$ associated with $p_i,p_k\in \mathcal{P}^3$.  

\par We set $\kappa^3_{p p_{dj}}=\max_{t \in [\mathcal{T}_{dj},\mathcal{T}_{d \; j+1}-1]} \kappa^2_{p p_{dt}}$ for all $p \in \{p_+,p_r \forall r \in \mathcal{R}\}$.  We set $\kappa^3_{ p_{+}p_{r}} = \kappa_{ p_{+}p_{r}}$.  

\par We set $\kappa^3_{ p_{dj}p_-}=\max_{t \in [\mathcal{T}_{dj},\mathcal{T}_{d \; j+1}-1]} \kappa^2_{p_{dt}p_-}$.  

\par For any pair of unique $d_i,d_k$ and windows $j_i,j_k$ \\ 
we set  $\kappa^3_{ p_{d_ij_i}p_{d_kj_k}}=\max_{\substack{t_0 \in [\mathcal{T}_{d_ij_i},\mathcal{T}_{d_i \; j_i+1}-1]\\ t_1 \in [\mathcal{T}_{d_kj_k},\mathcal{T}_{d_k \; j_k+1}-1]}} \kappa^2_{p_{d_it_0}p_{d_kt_1}}$. 

\par Evaluating each of the $\kappa^3$ terms amounts to solving a basic shortest path problem (no resource constraints), meaning not all $\kappa^2$ terms mentioned in these optimizations need be explicitly computed.  %The weight between a node $p \in \{p_+,p_r\}$ to $p_{dj}$ is the shortest path from $p_+$ or $p_r$ to $p_{dt}$ for $t \in [\mathcal{T}_{dj},\mathcal{T}_{d \; j+1}-1]$   or $\kappa^3_{p}=\min_{t}$

\par Replacing $\mathcal{E}^+$ with $\mathcal{E}^3$ in \eqref{objPath}-\eqref{resourceConst} we have a much smaller problem permitting more efficient optimization, which provides a lower bound on   \eqref{objPath}-\eqref{resourceConst}.  

Optimization produces a feasible route when each item in the route is associated with exactly one unique time. In pursuit of a feasible route, we add the times associated with items in the route to their respective $\mathcal{T}_d$ sets.  \par We iterate between solving the ERCSPP over $\mathcal{E}^3$ and augmenting the $\mathcal{T}_d$ until we obtain a feasible route.  This must ultimately occur since eventually $\mathcal{T}_d$ would include all $t \in \mathcal{T}$ for all $d \in \mathcal{D}$; though termination occurs much earlier in practice. 

We describe pricing formally below. We use $t_{p_{d_ij_i}p_{d_kj_k}0}$ and $t_{p_{d_ij_i}p_{d_kj_k}1}$ to denote the maximizers $ \argmax_{\substack{t_0 \in [\mathcal{T}_{d_ij_i},\mathcal{T}_{d_i \; j_i+1}-1]\\ t_1 \in [\mathcal{T}_{d_kj_k},\mathcal{T}_{d_k \; j_k+1}-1]}} \kappa^2_{p_{d_it_0}p_{d_kt_1}}$ \\
used to calculate $\kappa^3_{ p_{d_ij_i}p_{d_kj_k}}$. \par The term $t_{p_{d_ij_i}p_{d_kj_k}0}$ is the time component maximizer for $p_{d_ij_i}$ while $t_{p_{d_ij_i}p_{d_kj_k1}}$ is the time component maximizer for $p_{d_kj_k}$. These are the outgoing and incoming times, respectively, for the shortest path on $(\mathcal{P}^2,\mathcal{E}^2)$ between $p_{d_ij_i}$ and $p_{d_kj_k}$.  \par We use tot\_sz to keep track of the total number of elements in all $\mathcal{T}_d$ sets.  A growth in tot\_sz implies a mismatch between the incoming time and the outgoing time at an item location. In such cases, tot\_sz grows to narrow the time ranges for the sets, making it less likely to have a mismatch.  When tot\_sz does not grow, no mismatch occurred and the solution obtained represents a feasible route, therefore we terminate pricing.  \par Algorithm \ref{fastPricing} shows pseudocode for the pricing method described in Section \ref{removeExplicitTIme}.

\begin{algorithm}[!ht]%[1]%[H]
 \caption{Fast Pricing}
 %\jl{What are $t_{p_i,p_j,0}$ on line 8 and $t_{p_i,p_j,1}$ on line 12? I think they are undefined. }

\begin{algorithmic}[!ht] 
\State $\mathcal{T}_d \leftarrow [t_d^-,t_d^++1] \quad \forall d \in \mathcal{D}$
\label{init_sets}
\Repeat
\label{bigLoop}
\State tot\_size $\leftarrow \sum_{d \in \mathcal{D}}|\mathcal{T}_{d}|$
\label{getSz}
\State   $x\leftarrow $ Solve  Eq \eqref{objPath}-\eqref{resourceConst} over $\mathcal{E}^3$
%\State my_
\For{($p_{i},p_{k}) \in  \mathcal{E}^3$ s.t. $ x_{p_i,p_k}=1$ }
%\If{}
\If {$p_i \neq p_+$ and $p_i \neq p_r  $  for any $r \in \mathcal{R}$} 
\State Let $p_i$ correspond to item $d$
\State $\mathcal{T}_d \leftarrow \mathcal{T}_d \cup t_{p_i,p_k,0} $
\EndIf
\If {$p_k \neq p_-$ and $p_k \neq p_r  $  for any $r \in \mathcal{R}$} 
\State Let $p_k$ correspond to item $d$
\State $\mathcal{T}_d \leftarrow \mathcal{T}_d \cup t_{p_i,p_k,1} $
\EndIf
\EndFor
 \Until{ tot\_size=$\sum_{d \in \mathcal{D}}|\mathcal{T}_{d}|$}
\State Let $g$ correspond to the solution to \eqref{objPath} computed via optimization over $\mathcal{E}^3$
\State Return $g$
\end{algorithmic}
\label{fastPricing}
\end{algorithm}
\subsection{Partial Optimization of the Restricted Master Problem for Faster Pricing}
\label{partOptFaster}
Solving the pricing problem is the key bottleneck in computation experimentally.  One key time consumer in pricing is the computation of the $\kappa$ terms, which can easily be avoided by observing that $\kappa^2,\kappa^3$ terms are offset by changes in $\lambda_d$ and $\lambda_r$ but the actual route does not change so long as $\lambda_e$, $\lambda_p$, and $\lambda_t$ are fixed. We resolve the RMP fully only periodically so that we can perform several rounds of pricing using different $\lambda_d,\lambda_r$ terms leaving the $\lambda_e,\lambda_p,\lambda_t$ fixed. 
%\jl{I didn't understand this paragraph at all. $\kappa^2,\kappa^3$ are offset from what? Why does the actual route not change? What is the motivation of only updating some of the $\lambda$?}
\section{Heuristic Pricing}
\label{heurPRice}

To accelerate CG in operations research applications, one often solves the pricing problem approximately and efficiently instead of solving pricing exactly (\cite{danna2005branch,costa2019}). When heuristic pricing fails to produce a negative reduced cost column, an exact algorithm can then be called for pricing; thus in the end, the use of heuristic pricing need not lead to sacrificing the guarantees of exact inference. Heuristic pricing can also be used in problems where exact pricing is intractable but can be solved heuristically very well and efficiently (\citep{FlexDOIArticle}).

In this section we propose a heuristic pricing approach to solving the selection of the optimal route in $\mathcal{P}^3,\mathcal{E}^3$ that  exploits the following observation. Given any ordering of the item set $\mathcal{D}$, computing the lowest reduced cost column that is consistent with that ordering is computationally tractable. We express this mathematically as follows. 

\par We use $\mathcal{M}$ to denote the set of orderings of $\mathcal{D}$, which we index by $m$. 
\par We describe $m$ using $M^m_{d_id_j} \in \{0,1\}$ where $M^m_{d_id_j}=1$ IFF $d_i$ precedes $d_j$ directly or indirectly in ordering $m$. 
\par We use $\mathcal{G}_m$ to denote the subset of $\mathcal{G}$ consistent with ordering $m$.

The lowest reduced cost path evaluated on graph ($\mathcal{P}^3,\mathcal{E}^3$) respecting the ordering $m$ enforces the following
\begin{align}
\label{dynRecConst}
    (M^m_{d_1d_2}=0)\rightarrow (x_{p_ip_j}=0) \\ \quad  \forall p_i=(d_1 \in \mathcal{D},q_1 \in \mathcal{T}_{d_1}),p_j=(d_2 \in \mathcal{D},q_2 \in \mathcal{T}_{d_2}) \nonumber
\end{align}
Since \eqref{dynRecConst} enforces that no item is repeated in a route we can remove \eqref{resourceConst}  from consideration when solving $\min_{g \in \mathcal{G}_m}\bar{\Gamma}_g$. We now write $\min_{g \in \mathcal{G}_m}\bar{\Gamma}_g$ given $m$ as a polynomial time solvable dynamic program. 
\par Let us define $\mathcal{E}^{3m} $ as the subset of $ \mathcal{E}^3$ s.t. $(M^m_{d_1d_2}=1)$. Let us define for any given $p \in \mathcal{P}^3,c \in \{0,1,2..c_0\}$ the value $\rho_{pc}$ as the cost of the lowest cost path in $\mathcal{E}^{3m}$ starting at $p_+$ and ending at $p$, requiring exactly $c$ units of demand.

\par We define $\rho_{pc}$ recursively below.

\begin{align}
\label{myDynPRog}
    \rho_{p_-c} =\max_{(\bar{p},p) \in \mathcal{E}^{3m}} \kappa^3_{\bar{p}p}+\rho_{\bar{p}c} \quad \forall c \in \{0,1...c_0\} \\%\quad \quad p=p_-
    \rho_{p c}= \max_{(\bar{p},p) \in \mathcal{E}^{3m}} \kappa^3_{\bar{p}p}+\rho_{\bar{p}c-c_{d}}\quad  \forall  p=(d,j), c\in \{0,1... c_0-c_d\} \nonumber \\
    \rho_{p_+0}=0 \nonumber \\
    \rho_{p_rc_0-c_r}=\kappa_{p_+p_r} \nonumber 
\end{align}
%We compute the $\rho$ terms in order (first to last) following $m$.  Any $\rho$ term not defined by the iteration above has $\rho$ value of $\infty$.
%After solving the dynamic program in Eq \ref{myDynPRog} we can return one or more negative reduced cost columns produced. %
%
%As in Alg 2, since optimization in Eq \ref{myDynPRog} is defined on $(\mathcal{P}^3,\mathcal{E}^3)$, not $(\mathcal{P}^2,\mathcal{E}^2)$ then  Eq \ref{myDynPRog} may generate a route that does not agree on its incoming and outgoing times for a given item hence requiring that the partition of times 
Since each execution of \eqref{myDynPRog} is fast we solve \eqref{myDynPRog} using multiple different random orderings and retain the solution with lowest reduced cost.\\ 
\par We now study the probability that  our heuristic pricing algorithm produces the lowest reduced cost route when run multiple times. Specifically we show that given capacity $c_0$, and $n$ random orderings that we compute the lowest reduced cost route with probability $\phi$  where  $\phi \geq 1-(1-\frac{1}{c_0!})^n$.\\

\par We establish this as follows. Consider any random ordering of the items $m$ and route $g \in \mathcal{G}$. Since all orderings are equally likely then the probability that any given route $g$ containing $c_0$ items follows the ordering in $m$ is $\frac{1}{c_0!}$. Since the number of items in the lowest reduced cost route $g^*$ is no greater than $c_0$ then the probability that a random ordering supports $g^*$ is no less than $\frac{1}{c_0!}$. If we generate $n$ random orderings (with replacement) the probability that any one of those orderings is consistent with $g$ is 1- the probability that none are consistent with $g$ or $1-(1-\frac{1}{c_0!})^n$.  

\par In our experiments the units of demand at each pickup location can be $\{1,2,3\}$. The carrying capacity of each robot, $c_0$, is set to 6 units. \par If we run 25 rounds of heuristic pricing in each attempt to solve $\min_{g \in \mathcal{G}}\bar{\Gamma}_g$ in Algorithm \ref{fastPricing}, then we have a probability of finding the lowest reduced cost column which includes 1 through 6 stops (1,2,3,4,5,6), $\phi \geq$ [1, 1, 0.9858, 0.6549, 0.1888, 0.03415] respectively. \par It should be noted that since our heuristic algorithm employs a dynamic program, each ordering that is solved can produce multiple negative reduced cost routes. We also get multiple routes from solving over multiple orderings. CG convergence is often accelerated by returning multiple columns through each iteration. In practice we choose to return a designated number of columns from the union of negative reduced cost columns obtained over all orderings.

%Thus a dual optimal solution for which  heuristic pricing fails to generate a negative reduced cost column can be understood as satisfying an odd sort of dual optimal inequality which ensures that with various degrees of probability no column containing a given number of items of negative reduced cost exists.

\section{Experiments}
\label{exper}

We run two separate experiments to study our model. In the first one, we study the added value of our model, comparing it to a modified version employing MAPF. In the second one, we study the speedup obtained by employing our heuristic pricing algorithm. We evaluate on benchmark maps used in the MAPF literature and on synthetically generated maps.

For problems on benchmark maps, we generate instances by randomly assigning pickup items, and extant robot initial locations, to locations on the map.  % Extant robots also have their initial positions randomly assigned.
We set the launcher point to be a central position on the map. For synthetically generated maps, we generate problems by starting with an open, square grid and randomly positioning a set number of obstacles throughout the grid space. In these synthetic problems, items, extant robots, and the launcher have randomly assigned positions.% We assign the launcher position randomly.

If we require an exact solution to the ERCSPP during pricing, we do so using an exponential time dynamic program outlined in the next section (\ref{ERCSPP_SOlVER}). Similar to heuristic pricing, the algorithm is capable of returning multiple negative reduced cost columns, the optimal one being among them. We set the maximum number of columns delivered when using either heuristic or exact pricing as a problem parameter. We return the lowest reduced cost columns when more are available than can be returned.  

We update the Lagrangian multipliers $\lambda_t$, $\lambda_p$, $\lambda_e$, and the associated graph components every three CG iterations, unless we are unable to find a negative reduced cost column in a given iteration, in which case we update all dual variables and rerun pricing. If, immediately after all dual variables are optimized, pricing fails to find a negative reduced cost column, then we have finished optimization and conclude CG. To ensure feasibility for the initial round of CG, we initialize the RMP with a prohibitively high cost dummy route $g_{r,init}$ for each $r \in \mathcal{R}$, where all $G_{dg_{r,init}},G_{tg_{r,init}},G_{pg_{r,init}},G_{eg_{r,init}}=0$ but $G_{rg_{r,init}}=1$.  These dummy routes represent an active robot route for each $r \in \mathcal{R}$, and thus guarantee that \eqref{max_rob} is satisfied. These extra robots ensure feasibility, but are not active at termination of CG due to their prohibitively high cost. Experiments are run in MATLAB, and CPLEX is used as our general purpose (mixed integer) linear programming solver. Our machine is equipped with a 8-core AMD Ryzen 1700 CPU @3.0 GHz and 32 GB of memory running Windows 10.

\begin{figure}[!hbtp]
    \centering
	\includegraphics[width=0.25\linewidth]{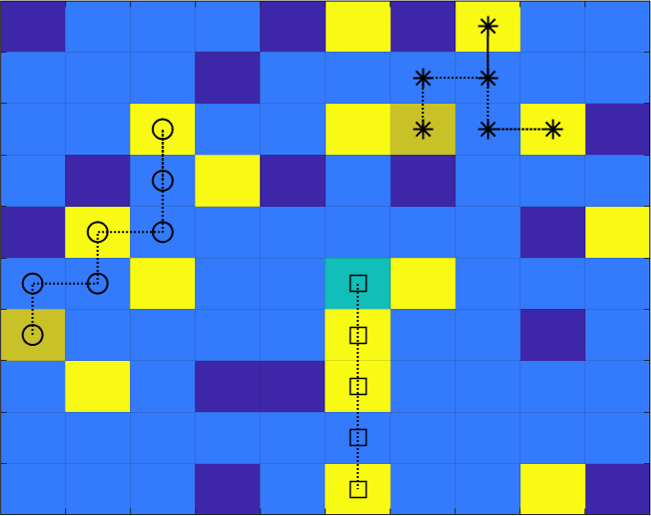}
	\includegraphics[width=0.25\linewidth]{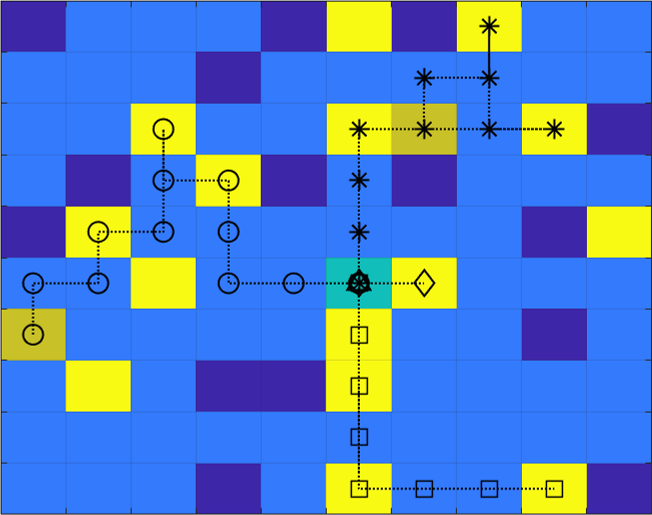}
    \includegraphics[width=0.25\linewidth]{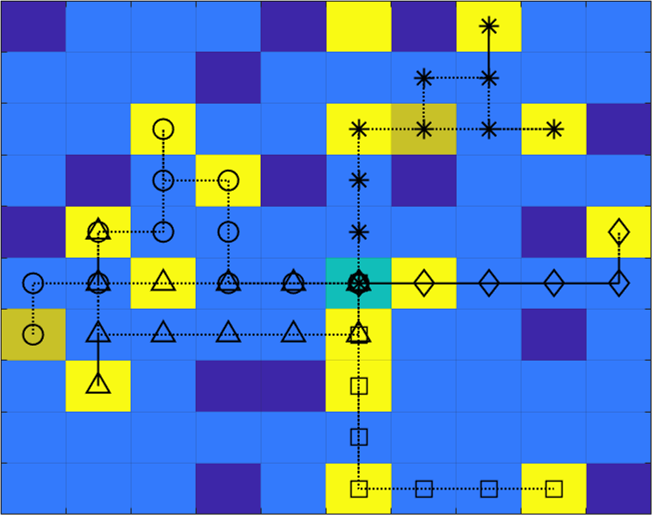} \\
    \includegraphics[width=0.65\linewidth]{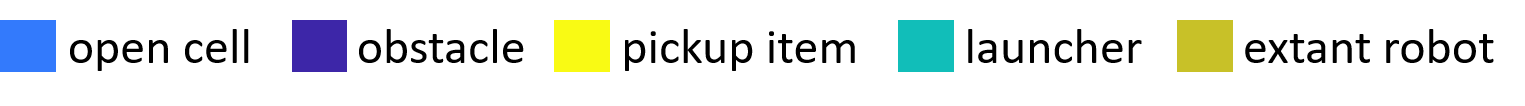}
	\caption{Robot route results for a single instance over 3 snapshots in time.  Each track is a robot route up through that time step.  Traversable cells, obstacles, the starting/ending launcher, item locations, and extant robot locations are all noted in the legend.\\
		\textbf{(Left): t = 8 snapshot}\\   
		\textbf{(Middle): t = 16 snapshot}\\
		\textbf{(Right): t = 30 (end time) snapshot}}
	\label{fig:paths}
\end{figure}

%\begin{comment}
A sample problem with the solution routes is shown in Figure \ref{fig:paths}.  Each plot in the Figure \ref{fig:paths} shows a snapshot in time of the same instance's solution.  A snapshot shows each robot's route from the initial time up to the time of the snapshot.

%\end{comment}

\subsection{Elementary Resource-Constrained Shortest-Path Solver}
\label{ERCSPP_SOlVER}
We solve the elementary resource-constrained shortest-path problem (ERCSPP) in pricing via an exponential time dynamic program that iterates over the possible remaining capacity levels for a robot (starting at the highest), enumerating all available routes corresponding to paths in $(\mathcal{P}^3,\mathcal{E}^3)$ at each capacity level, and then progressing to the next highest remaining capacity level.  At each level, we eliminate any inferior routes.  We call a route inferior to another IFF all of the following are satisfied: (1) it has the same remaining capacity and corresponding position in the node set $\mathcal{P}^3$ as the other, (2) it has higher cumulative edge profit on $(\mathcal{P}^3,\mathcal{E}^3)$ than the other, and (3) it has a set of serviceable items available to it that is a subset of the other's.

We start at the maximum robot capacity and enumerate all possible single visit traversals.  We save a robot state for each such route.  A robot state is defined by its current corresponding position in the node set $\mathcal{P}^3$, the items serviced, the cost incurred so far on $(\mathcal{P}^3,\mathcal{E}^3)$, and the remaining capacity.  We set $\mathcal{K}_{p,h}$ to be the cost of a path at graph position $p \in \mathcal{P}^3$ with path history $h$, a set of all previously visited graph positions.  We set $\mathcal{C}_{p,h}$ to be the remaining capacity available for a robot at corresponding graph position $p$ with history $h$.  For a robot route with initial visit at item $d$ at corresponding graph position $p_{dj}$ we have the following remaining capacity and cost.

\begin{align}
\label{dp:cost_update}
    \mathcal{K}_{p_{dj},\{p_+\}} = \kappa^3_{ p_+p_{dj}}\\
    \mathcal{C}_{p_{dj},\{p_+\}} = c_0-c_d
\end{align}

We then move on to the next highest remaining robot capacity level.  For each saved robot state at this remaining capacity, we enumerate all available single visit traversals (including back to the launcher) and save a state for each route generated.  An item is available to be visited if that item has not yet been visited in the route and visiting it would not exceed the remaining capacity.  For a robot traveling from corresponding graph position $p_{d_ij_i}$ with history $h$, to corresponding graph position $p_{d_kj_k}$, we have the following update for the cost and remaining capacity.

\begin{align}
\label{dp:cost_update2}
    \mathcal{K}_{p_{d_kj_k},h\cup p_{d_ij_i} } = \mathcal{K}_{p_{d_ij_i},h} + \kappa^3_{ p_{d_ij_i}p_{d_kj_k}}\\
    \mathcal{C}_{p_{d_kj_k},h\cup p_{d_ij_i}} = \mathcal{C}_{p_{d_ij_i},h} - c_{d_k}
\end{align}

We eliminate all inferior routes generated and continue on to the next capacity level until we have exhausted all possible remaining capacity levels.  At the end, we have series of routes drawn out, including the route with minimum cost on $(\mathcal{P}^3,\mathcal{E}^3)$.  We can return any number of these that have a negative cost.  Returning more serves to reduce the number of CG iterations, but comes with a trade-off of burdening the RMP with more, possibly unnecessary, columns. % Ultimately, we choose to return the twenty lowest reduced cost routes found.

\subsection{Comparison with MAPF}

We compare our algorithm to a modified version that incorporates MAPF. This version will initially ignore robot collision constraints but ultimately considers them after a set of serviceable items are assigned to specific robots. The modified algorithm works as follows. We solve a given problem instance using our CG algorithm, but we neglect the collision constraints, meaning $\lambda_p=0,\lambda_e=0, \forall p\in \mathcal{P}, e\in \mathcal{E}$ and that the IP solved is defined by equations (1)-(4). This is a vehicle routing problem which delivers us a set of time-window feasible robot routes, including the items serviced by each robot; however, these routes could include collisions. We then take the disjoint set of items serviced (routes) and feed them to a MAPF solver \citep{LiAAMAS20b} employing Priority Based Search \citep{ma2019searching}. The MAPF solver delivers a set of non-colliding robot routes, each attempting to service the set of items assigned to it. If the MAPF solver fails to provide a valid route for a particular robot (i.e., it cannot make it back to the launcher in time) that route is excluded in the algorithm's final solution. Since standard MAPF solvers can not handle time windows on items we ignore the time windows for the MAPF solver, but not our CG solver. This provides an advantage to the MAPF solver by permitting it to use routes that do not obey time windows associated with items. When we say permitting here, we need to clarify. The ordered disjoint sets are time-window feasible, but by eliminating collisions (causing robots to wait enroute) we could produce routes which are then time-window infeasible (but not by much since the original routes are time-window feasible). We do not worry about this issue -- only that the robots return to the launcher by the end of the time period. %By this we mean the the MAPF solver  in terms of the space of feasible solutions. 

\begin{figure}[!htbp]
    \centering
	\includegraphics[width=.7\linewidth]{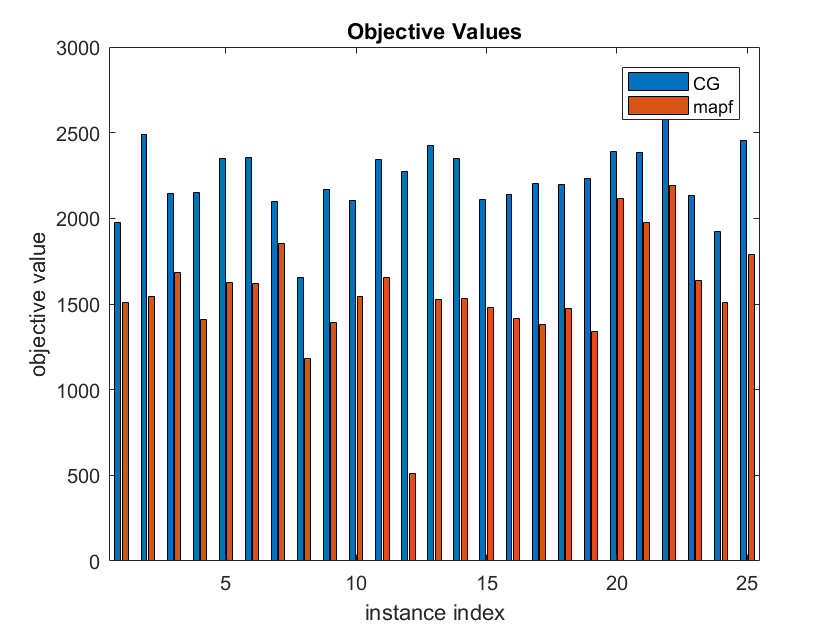}
	
	\caption{Objective values for both approaches over each problem instance. Our full CG approach is shown in blue.  It is compared against the modified column CG + MAPF approach shown in orange.}
	\label{fig:mapf_comparison}
\end{figure}

We compare the resulting objective values from our full CG approach to this modified approach. We solve 25 instances on the 32x32 grid \texttt{maze-32-32-2} that was presented recently in \citep{SternSoCS19}. Each problem instance has 60 pickup items (of size 1-3), 8 total robots, 2 extant robots, and 150 total time steps. We set $\theta_{1}$ to -1, $\theta_{2}$ to -1, and the reward for servicing any item, $\theta_{d}$, to 100. Each robot, including the extant ones, has a capacity of 6, while each item has a random size (capacity consumption) uniformly distributed over the set \{1,2,3\} units. Note that this choice is arbitrary. We could also set the capacity of each extant robot to be  random integer between 1 and 6. In each round of pricing we return the 50 lowest reduced cost columns found. Each item's time window is randomly set uniformly over the available times and can be up to 50 time periods wide. %For the purposes of this comparison however, we neglect time constraints for the items specifically when employing MAPF so as to be generous to the MAPF solver, which is not equipped to handle time windows for items.
We compare our final results with time windows to the MAPF algorithm's results without them. The objective value results for both approaches are show in table \ref{table:mapf_comparison}. A side by side plot of the objective values is shown in Figure \ref{fig:mapf_comparison}.

\begin{table}[h!]
	\centering
	\scalebox{0.7}{
		\begin{tabular}{|c|c|c|c|c|} 
		\hline
		    & \bf CG & \bf modified CG + MAPF & \bf Difference (CG - MAPF) & \bf Relative Difference \\
		\hline
		\bf mean & 2230.1 & 1555.5 & 674.6 & 30.21\\
		\hline
		\bf median & 2202.0 & 1535.0 & 685.0 & 31.10 \\
		\hline
	\end{tabular}}
	\caption{Objective value results for both algorithms over 25 random instances.  Our full approach is labeled CG.  We compare against modified CG + MAPF.}
	\label{table:mapf_comparison}
\end{table}

We see an average objective difference of 921 revenue units and a median difference of 782 revenue units from the modified algorithm to our full algorithm.  We note from looking at Figure \ref{fig:mapf_comparison} that each of the 25 instances show drastic improvements for our algorithm.  These instances largely include robot routes that the MAPF algorithm was unable to find a complete route for within the time constraint given the potential collisions with other robots.  With such problems we see it is critical to employ our full algorithm that jointly considers routing and assignment.

Runtime results, iteration counts, and objective values for our full CG approach on the 25 problem instances are shown in Table \ref{table:runtime_20}. We also look at the LP objective of the CG solution and the corresponding relative gaps. The relative gap is defined as the the absolute difference between the upper bound (the LP objective value) and our integer solution (the lower bound) divided  by the upper bound. We normalize so as to efficiently compare the gap obtained (upper bound - lower bound) across varying problem instances. Please note that the runtimes are of theoretical interest only. In practice we would use the heuristic pricing speedup discussed in the next section. 

\begin{table}[t!]
	\centering
	\scalebox{0.6}{
		\begin{tabular}{|c|c|c|c|c|c|}
		%\hline
		%\multicolumn{6}{|c|}{\bf 20x20 grid Results } \\
		\hline
		\bf  & \bf Time (sec) & \bf Iterations & \bf LP Objective & \bf Integral Objective & \bf Relative Gap \\
		\hline
		\bf mean & 17577.5 & 65.9 & 2347.6 & 2230.1 & .05 \\
		\hline
		\bf median & 6526.1 & 67 & 2289.8 & 2202.0 & .05 \\
		\hline
	\end{tabular}}
	\caption{Results of the full CG approach over 25 problem instances.}
	\label{table:runtime_20}
\end{table}

\subsection{Heuristic Pricing Speedup}

\begin{figure}[!htbp]
    \centering
	\includegraphics[width=.7\linewidth]{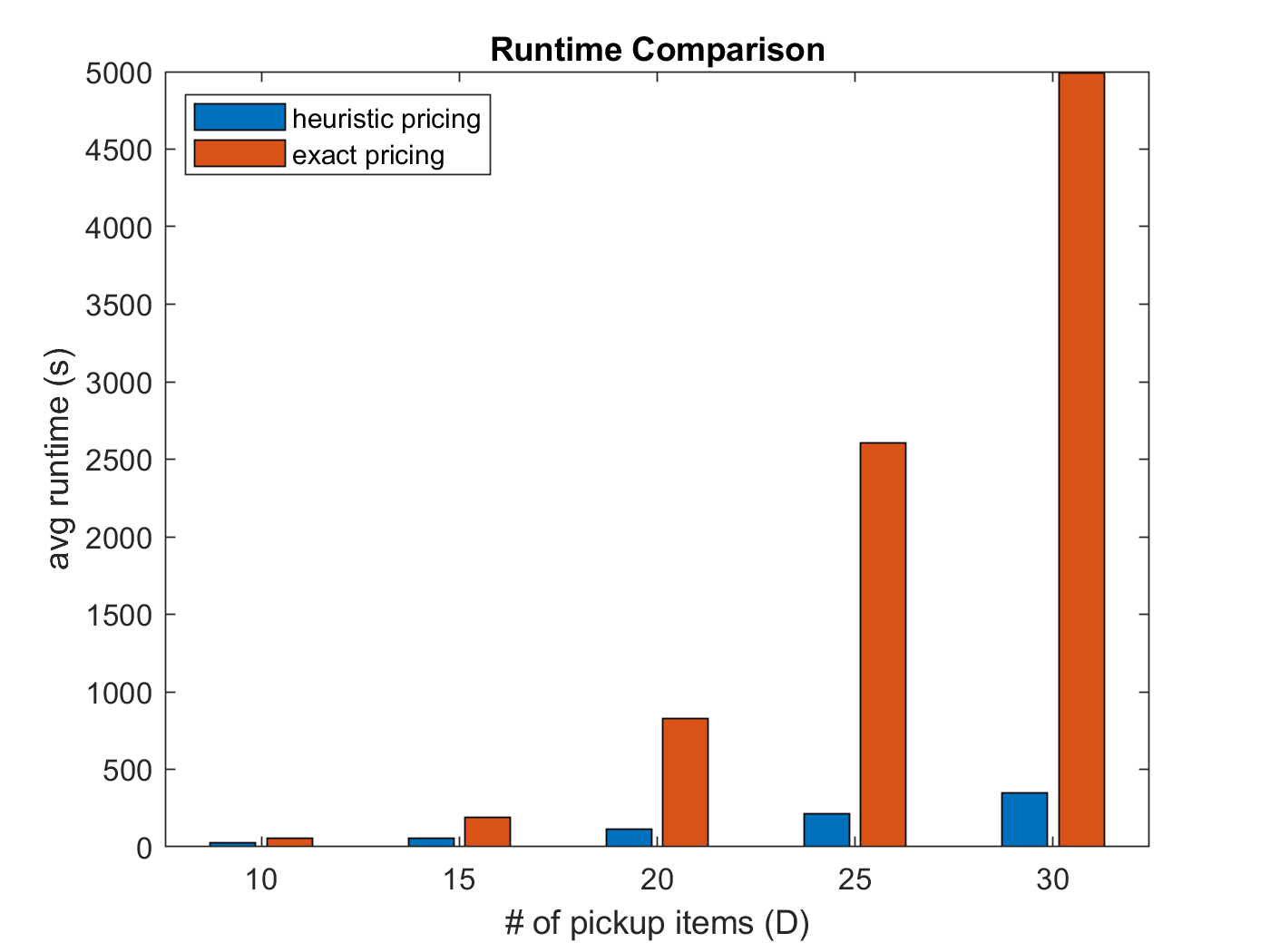}
	\caption{Average runtime results over problems with increasing numbers of pickup items. Runtimes were averaged over 10 random instances.}
	\label{fig:runtime_comparison}
\end{figure}

There is no question that without further improvements, the CG solver is too slow to work in practice. However we find that our heuristic pricing yields considerable speedup. We run experiments to measure the speedup offered by our heuristic pricing solver. We compare two approaches. In the first approach, we employ heuristic pricing in each iteration but ultimately employ exact pricing if heuristic pricing fails in any iteration. In this scenario, exact pricing must be employed at least once in order to ensure optimality of the LP solution. In the second approach, we employ exact pricing at each iteration. We solve on random problem instances with randomly generated grids. Each experiment is run on a 25x25 grid with 50 random obstacles, 5 total robots, 2 extant robots, and 75 time steps. Robots have a capacity of 6 while each pickup item has a uniform random demand in the set $\{1,2,3\}$. $\theta_d$ is set to 100 while $\theta_1$ and $\theta_2$ are both set to -1. Each item's time window is randomly set uniformly over the available times and can be up to 25 time periods wide. We return the lowest 25 reduced cost columns found when employing heuristic or exact pricing. We run this problem setup for different pickup item counts ranging from 10 to 30 in increments of 5. For each pickup item count, we run 10 random instances and record the average runtime over the instances. Numerical results are shown in Table \ref{table:runtime_comparison} and a corresponding plot is shown in Figure \ref{fig:runtime_comparison}.

\begin{table}[h!]
	\centering
	\scalebox{0.7}{
		%\begin{tabular}{c|c|c|c} 
		\begin{tabular}{lrrr}
		   \bf D & \bf EP & \bf HP & \bf speedup (x) \\
		   \hline
		\bf 10 & 56.1 & 25.5 & 2.1\\
		\bf 15 & 192.2 & 55.7 & 3.4\\
		\bf 20 & 826.8 & 114.4 & 6.8\\
		\bf 25 & 2605.9 & 211.8 & 10.7\\
		\bf 30 & 4989.0 & 346.1 & 13.1\\
		\hline
	\end{tabular}}
	\caption{Average runtime results in seconds over problems with various numbers of pickup items when using exact pricing (EP) and when using heuristic pricing (HP). Runtimes were averaged over 10 random instances.}
	\label{table:runtime_comparison}
\end{table}

It can be observed that employing heuristic pricing offers a positive average speedup for all pickup item counts. This speedup starts small for 10 pickup items, but grows considerably as the number of pickup items is increased. We see that the real value in the heuristic pricing solver is its scalability in comparison to exact pricing. Further speedups would be considered in a practical application of this method, and taken together these would yield impressive time savings. See for example the work of \citep{desaulniers2002accelerating}.

\section{Conclusion and Future Research}
\label{conc}
In this paper, we unified the work on multi-agent path finding with the vehicle routing-column generation literature to produce a novel approach applicable to broad classes of Multi-Robot Routing (MRR) problems. The new decade accelerated an already rapid transformation of many warehouse operations to automation, thereby increasing the importance of these problems as a key factor in agile supply chains. Our work treats MRR as a weighted set packing problem where sets correspond to valid robot routes and elements correspond to space-time positions. Pricing is treated as an elementary resource-constrained shortest-path problem (ERCSPP), which is NP-hard, but solvable in practice \citep{irnich2005shortest}. We solve the ERCSPP by adapting the approach of \citep{boland2017continuous} to limit the time windows that need be explored during pricing. We introduce a heuristic pricing algorithm to efficiently solve the ERCSPP problem. While this speeds up the processing considerably and probably results in a model which can be implemented on a rolling-horizon basis (with heuristic real-time insertion techniques), further improvement in the pricing problem would be helpful. Our ongoing research shows great promise in this regard. 

The easiest future work is to transform this formulation to the closely related problem in which robots collect pallets and move these to staging areas where human or automate pickers select items. These pallets are then returned to the warehouse floor. That version of the MRR problem can easily be solved within our framework. 
\par Our next future task will be to tighten the LP relaxation using subset-row inequalities \citep{jepsen2008subset} and ensure integrality with branch-and-price \citep{barnprice}. Subset row inequalities are trivially applied to sets over the pickup items since they do not alter the solution paths. Similarly, branch-and-price could be applied to sets over pickup items, following the vehicle routing literature  \citep{Desrochers1992}.  %As well, we intend to incorporate ng-route \citep{baldacci2011new} relaxations which provide polynomial time pricing at the expense of loosening the master problem relaxation. Specifically, ng-route relaxations permit routes to contain multiple copies of items as long as a distant (in space) item is visited between copies of a given item.
We also seek to provide insight into the structure of dual optimal solutions and study the effect of smoothing in the dual, based on the ideas of \citep{haghani2020smooth,haghani2020relaxed}. Simply put, we suspect that dual values change smoothly across space and time, thus we will encourage such solutions over the course of column generation. Finally, we intend to apply our column generation solution techniques to a wide variety of transportation and logistics optimization problems.

\bibliographystyle{abbrvnat} % outcomment this and next line in Case 1
\bibliography{col_gen_bib}

%\section{Figure Candidates}
\end{document}